\documentclass[11pt]{article}
\usepackage{acl-hlt2011}
\usepackage{times}
\usepackage{latexsym}
\usepackage{amsmath}
\usepackage{multirow}
\usepackage{url}
\usepackage{graphics}
\usepackage{graphicx}
\newenvironment{myitemize}{
\vspace{-0.6\baselineskip}
\begin{itemize}
\setlength{\topsep}{0pt}
\setlength{\itemsep}{3pt}
\setlength{\parskip}{0pt}
\setlength{\parsep}{0pt}
\setlength{\partopsep}{0pt}
}{
\end{itemize}
\vspace{-0.4\baselineskip}}
\newenvironment{myenumerate}{
\vspace{-0.3\baselineskip}
\begin{enumerate}
\setlength{\topsep}{0pt}
\setlength{\itemsep}{1pt}
\setlength{\parskip}{0pt}
\setlength{\parsep}{0pt}
\setlength{\partopsep}{0pt}
}{
\end{enumerate}
\vspace{-0.2\baselineskip}}

 \title{Tracking Sentiment in Mail:\\ How Genders Differ on Emotional Axes}
 \author{Saif M. Mohammad$^\dagger$ \and Tony (Wenda) Yang$^{\dagger\star}$\\
     Institute for Information Technology,
     National Research Council Canada$^\dagger$.\\
     Ottawa, Ontario, Canada, K1A 0R6.\\
 	School of Computing Science,
 	Simon Fraser University$^\star$\\
 	Burnaby, British Columbia, V5A 1S6.\\
     {\tt saif.mohammad@nrc-cnrc.gc.ca, wenday@sfu.ca}
 }

\date{}

\begin{document}
\maketitle
\begin{abstract}
% Emotions are central to our well-being, yet it is hard to
% be objective one's own emotional state.
% However, 
% How we feel toward different people and entities
% is expected to manifest itself
% in what we speak and write. 
% Letters have long been a channel to convey emotions, explicitly and implicitly.
With the widespread use of email,
we now have access to unprecedented  amounts of text that we ourselves
have written.
In this paper, we show how 
sentiment analysis can be used in tandem with effective
visualizations to quantify and track emotions in many types of mail.
We create a large word--emotion association lexicon by 
crowdsourcing, and use it to
compare emotions in love letters, hate mail,
and suicide notes. 
% Next we analyze the difference in emotion words used by men and women in email. 
We show that there are marked differences across genders
in how they use emotion words in work-place email. For example, women use many
words from the joy--sadness axis, whereas men prefer terms
from the fear--trust axis.
Finally, we show
visualizations that can help people track emotions in their
emails.
% We will make our emotion lexicon publicly available, as well as the love letter corpus, the hate mail corpus,
% and the suicide notes corpus.
%  We present an online emotion analyzer, that given an input
%  text, quantifies the emotions present in it. It generates
% visualizations aimed at: (1) quickly and effectively
% conveying this information to the user, (2) showing how
% emotions associated with a target entity change over time,
% and (3) bringing to attention deviations from norm.  The
% visualizer has a wide range of applications including
% social and literary analysis of books, newspapers, letters, and social
% network information. It can even be used to monitor a continuous
% stream of text to detect spikes and dips in emotions
%% associated with an entity of interest, say a commercial
% product or a presidential candidate.  
\end{abstract}
\section{Introduction}
Emotions are central to our well-being, yet it is hard to
be objective of one's own emotional state.
% However, what we feel toward different people and entities is expected to manifest itself in what we speak and write. 
Letters have long been a
channel to convey emotions, explicitly and implicitly,
and now with the widespread usage of email,
people have access to unprecedented amounts of text that they
themselves have written and received.
In this paper, we show how  
sentiment analysis can be used in tandem with effective
visualizations to track emotions in letters and emails.

Automatic analysis and tracking of emotions in emails has a number of benefits including:
% \vspace*{-2mm}
\begin{myenumerate}
\item Determining risk of repeat attempts by analyzing suicide notes \cite{Osgood59,Matykiewicz09,Pestian08}.\footnote{The 2011 Informatics for Integrating Biology and the Bedside (i2b2) challenge by 
the National Center for Biomedical Computing is on detecting emotions in suicide notes.}
\item Understanding how genders communicate through work-place and personal email \cite{Boneva01}.
\item Tracking emotions towards people
and entities, over time. 
For example, did a certain managerial course bring about a measurable change
in one's inter-personal communication?
\item Determining if there
is a correlation between the emotional content of letters and changes in a person's social, economic, or physiological state.
Sudden and persistent changes in the amount of emotion words in mail may be
a sign of psychological disorder.
\item Enabling affect-based search. For example, efforts to improve customer satisfaction can benefit by searching the received mail for snippets expressing anger \cite{DiazR02,DubeM96}. 
\item Assisting in writing emails that convey only the
desired emotion, and avoiding misinterpretation \cite{LiuLS03}.
% \item Determining if effective orators and writers make more use of emotive words than others.
\item Analyzing emotion words and their role in persuasion in communications by fervent letter writers such as Francois-Marie Arouet Voltaire and Karl Marx 
\cite{Voltaire73,Marx82}.\footnote{Voltaire: http://www.whitman.edu/VSA/letters\\
\indent $\;\;$ Marx: http://www.marxists.org/archive/marx/works/date}
% \item Gaining insights into how society has changed over time. For example, tracking how letters and email portray women.
\end{myenumerate}
\vspace*{-2mm}
\noindent In this paper, 
we describe how we created  a large word--emotion association lexicon by crowdsourcing with effective quality control measures (Section 3).
% We show how we created a simple emotion tracker that relies on the powerful emotion lexicon.
In Section 4, we show comparative analyses of emotion words in love letters, hate mail,
and suicide notes. This is done: (a)  To determine the distribution of emotion words in these types of mail,
as a first step towards more sophisticated emotion analysis (for example, in developing a depression--happiness scale for Application 1),
and (b) To use these corpora as a testbed to establish that the emotion lexicon
and the visualizations we propose help interpret the emotions in text.
% as a sanity check of the manually created emotion
% lexicon, (b) to better understand the composition of hate mail and suicide notes.\footnote{ a preliminary step before
% pursuing our future goal of creating a happiness--depression scale (Application 1 from the previous page).
In Section 5, we analyze how men and women differ in the kinds of emotion words they use in
work-place email (Application 2).
Finally, in Section 6, we show how emotion analysis can be integrated with email
services such as Gmail to 
help people track emotions in the
emails they send and receive (Application 3).
% We created the word--emotion association lexicon by crowdsourcing with effective
% quality control measures.
% Given an input text, the analyzer quantifies the emotions in it by simply
% counting the number of different emotion words in it. Next it generates
% visualizations aimed at (1) quickly and effectively conveying the emotion information, 
% (2) showing how emotions associated with a target entity change over time, and
% (3) bringing to attention deviations from norm. % , and, more generally, providing insights into the data. 
% % The emotion analyzer quantifies the emotions present in a piece of text, quantifies how the strengths of emotions in one piece of text
% % deviate from the strengths in another, and how emotions associated with a target entity change over time.
% We use the emotion analyzer to first we compare emotion words in love letters, hate mail,
% and suicide notes. Then we investigate the differences in the use of emotion words
% between men and women by analyzing the {\it Enron email Corpus} \cite{Enron}.\footnote{The Enron email corpus
% is available at http://www-2.cs.cmu.edu/~enron}

The emotion analyzer recognizes words with positive polarity (expressing a favorable sentiment towards an entity), negative polarity (expressing an unfavorable sentiment towards an entity), and no polarity (neutral). 
It also  associates words with joy, sadness, anger, fear,
trust, disgust, surprise, anticipation, which are argued to be the eight
basic and prototypical emotions \cite{Plutchik80}. 
% We now present details about the emotion lexicon and how it is
% used to analyze text (Section 2), followed by the various visualizations and applications they are suited for (Section 4),
% and lastly we present future work (Section 5).

  \section{Related work}
Over the last decade, there has been considerable work in sentiment analysis, especially in determining whether a term has a positive
or negative  polarity \cite{Lehrer74,Turney03,MohammadDD09}. 
% (This sense of {\it polarity} is also referred to as {\it semantic orientation}.)
There is also work in more sophisticated aspects of sentiment, for example, in detecting emotions such as
anger, joy, sadness, fear, surprise, and disgust \cite{Bellegarda10,MohammadT10,AlmRS05,AlmRS05}.
The technology is still developing and it can be unpredictable when dealing with short sentences,
but it has been shown to be reliable when drawing conclusions from large amounts of text \cite{DoddsD10,PangL08}.
%(see Pang and Lee \shortcite{PangL08} for a detailed review of relevant research). 

  Automatically analyzing affect in emails has primarily been done for automatic gender identification \cite{EnronGender,Corney02}, but it has relied on
  mostly on surface features such as exclamations and very small emotion lexicons. % (a few hundred words).
% There are a few different word--emotion lexicons available, but they tend to have a very small coverage.
The WordNet Affect Lexicon (WAL) \cite{StrapparavaV04} has a few hundred words
annotated with associations to a number of affect categories including the six Ekman emotions (joy, sadness, anger, fear,
disgust, and surprise).\footnote{WAL: http://wndomains.fbk.eu/wnaffect.html} 
% by manually identifying the emotions of a few seed words and then
% marking all their WordNet synonyms as having the same emotion.
% The words in WAL are annotated for a number of emotion and affect categories, but its creators
% also provided a subset corresponding to the six Ekman emotions.
General Inquirer (GI) \cite{Stone66} has 11,788 words labeled with 182
categories of word tags, including positive and negative polarity.\footnote{GI: http://www.wjh.harvard.edu/$\sim$inquirer} 
% It has certain other affect categories too, such as arousal and pain, but not categories corresponding to the basic emotions proposed by Ekman. 
Affective Norms for English Words (ANEW) has pleasure (happy--unhappy), arousal (excited--calm), and dominance (controlled--in control) ratings
for 1034 words.\footnote{ANEW: http://csea.phhp.ufl.edu/media/anewmessage.html}
Mohammad and Turney \shortcite{MohammadT10} compiled emotion annotations
for about 4000 words with eight emotions (six of Ekman, trust, and anticipation). % , but their lexicon is not publicly available.
% We created a word--emotion lexicon for over 10,000 words and the eight Plutchik emotions.

\section{Emotion Analysis}
% We now describe how we created a large word--emotion association lexicon (the {\it emotion lexicon}), and a simple word-counting method to analyze text.

\subsection{Emotion Lexicon}
% -- Crowdsourcing the emolex.\\
% -- Lexicon numbers

We created a large word--emotion association lexicon by crowdsourcing to Amazon's mechanical Turk.\footnote{Mechanical Turk:
www.mturk.com/mturk/welcome} We follow the method outlined in Mohammad and Turney \shortcite{MohammadT10}. Unlike Mohammad and Turney, who used the {\it Macquarie Thesaurus} \cite{Bernard86},
we use the {\it Roget Thesaurus} as the source for target terms.\footnote{Macquarie Thesaurus: www.macquarieonline.com.au\\ 
% to be annotated with associated emotions.\footnote{Macquarie Thesaurus: www.macquarieonline.com.au\\
\indent $\;\;$ Roget's Thesaurus: www.gutenberg.org/ebooks/10681}  
Since the 1911 US edition of {\it Roget's} is available freely in the public domain, it allows us
to distribute our emotion lexicon without the burden of restrictive licenses. We annotated only those words that occurred more than
120,000 times in the Google n-gram corpus.\footnote{The Google n-gram corpus is available through the LDC.} %Linguistic Data Consortium.}
% The frequency threshold of 120,000 is somewhat arbitrary. 
% We simply wanted to annotate the most frequent words first.}

 The {\it Roget's Thesaurus} groups related words into about a thousand categories,
%  Each category has a {\it head word} that best represents the words in it.
 which can be thought of as coarse senses or concepts \cite{Yarowsky92}.
 If a word is ambiguous, then it is listed in more than one category.
 Since a word may have different emotion associations when used in different senses,
 we obtained annotations at word-sense level by first asking 
an automatically
 % We were additionally interested in determining colour signatures for emotions (Section 7),
generated word-choice question pertaining to the target:

 \vspace*{2mm}
{\small
\noindent Q1. Which word is closest in meaning  % (most related) 
to {\it shark} (target)?

 \vspace*{1mm}
 \begin{minipage}[t]{4mm}
 \end{minipage}
 \begin{minipage}[t]{1.5cm}
  \begin{myitemize}
 \item {\it car}
 \end{myitemize}
 \end{minipage}
 \begin{minipage}[t]{1.5cm}
  \begin{myitemize}
 \item {\it tree}
 \end{myitemize}
 \end{minipage}
 \begin{minipage}[t]{1.5cm}
  \begin{myitemize}
 \item {\it fish}
 \end{myitemize}
 \end{minipage}
 \begin{minipage}[t]{2.2cm}
  \begin{myitemize}
 \item {\it olive}
 \end{myitemize}
 \end{minipage}
}
\noindent The near-synonym is taken from the thesaurus, and the distractors are randomly chosen words.
This question guides the annotator to the desired sense of the target word.
It is followed by ten questions asking if the target  is associated with positive sentiment, negative sentiment,
 anger, fear, joy, sadness, disgust, surprise, trust, and anticipation. The questions are phrased exactly
 as described in Mohammad and Turney \shortcite{MohammadT10}. 

% The near-synonym also guides the annotator to the desired sense of the word.
% Further, it encourages the annotator to think clearly about the target word's meaning; we believe this improves the quality of the annotations in Q2.
%  If a word has multiple senses, that is, it is listed in more than one thesaurus category,
%  then separate questionnaires are generated for each sense.
%  Thus we obtain colour associations at a word-sense level.
% Thesauri such as the Roget and Macquarie divide the vocabulary into about a thousand coarse categories. Words in a category are closely related, and an ambiguous word may be listed
% in more than one category. These categories can be thought of as its coarse senses. Like Mohammad and Turney \shortcite{MohammadT10}, our emotion
% lexicon has entries at sense level. We bias the annotator towards a particular sense of the target word by presenting a related word from the
% target thesaurus category (target sense).
% For each target word, annotators were asked (through ten separate questions) if it is associated with positive sentiment, negative sentiment,
% anger, fear, joy, sadness, disgust, surprise, trust, and anticipation.
If an annotator answers Q1 incorrectly,
then we discard information obtained from the remaining questions.
Thus, even though we do not have correct answers to the emotion association questions, likely incorrect annotations are filtered out.
About 10\% of the annotations were discarded because of an incorrect response to Q1.

% Each term has, on average, data from 5.1 distinct annotators, and most have exactly five.
Each term is annotated by 5 different people.
For 74.4\% of the % those 
instances, all five annotators agreed on whether a term is associated
with a particular emotion or not. For 16.9\% of the instances four out of five people agreed with each other.
% and for 8.5\% of the instances there was a three--two split.
The information from multiple annotators for a particular term is combined
by taking the majority vote. 
The lexicon has entries for about 24,200 word--sense pairs. The information from
different senses of a word is combined 
by taking the union of all emotions associated with the different senses of the word.
This resulted in a word-level emotion association lexicon for about 14,200 word types.
These files are together referred to as the {\it NRC Emotion Lexicon version 0.92}.

\subsection{Text Analysis}
Given a target text, the system determines which of the words exist in our emotion lexicon
and calculates ratios such as the number of words associated with an emotion to the total number of emotion words 
in the text. This simple approach may not be reliable in determining if a particular
sentence is expressing a certain emotion, but it is reliable in determining if a large piece of text
has more emotional expressions compared to others in a corpus.
% This is similar to the method employed by Dodds and Danforth \shortcite{DoddsD10}, who used a lexicon an order of magnitude smaller than ours.
Example applications include detecting spikes in anger words in close proximity to
mentions of a target product in a twitter stream \cite{DiazR02,DubeM96}, 
and literary analyses of text, for example, how novels and fairy tales differ in the use of emotion words \cite{Mohammad11b}.

\section{Love letters, hate mail, and suicide notes}

In this section, we quantitatively compare the emotion words in love letters,
hate mail, and suicide notes. 
% This preliminary investigation 
% has the following objectives: (1) To compile datasets pertaining to these three
% types of letters, (2) To determine the distribution of emotion words in these corpora
% as a first step towards more sophisticated emotion analysis,\footnote{For example, developing a depression--happiness scale.}
% and (3) To use these corpora as a testbed to establish that the emotion lexicon
% and the visualizations we propose help interpret the emotions in text.
We compiled a {\it love letters corpus (LLC) v$\,$0.1} by extracting 348 postings from lovingyou.com.\footnote{LLC: http://www.lovingyou.com/content/inspiration/\\loveletters-topic.php?ID=loveyou}
We created a {\it hate mail corpus (HMC) v$\,$0.1} by collecting 279 pieces of hate mail sent to the {\it Millenium Project}.\footnote{HMC: http://www.ratbags.com} 
%---a website that claims to highlight other websites that ``don't deserve a place on the web".
% We do not, in anyway, subscribe to the agenda of the website
The {\it suicide notes corpus (SNC) v$\,$0.1} has 21 notes taken from Art Kleiner's website.\footnote{SNC: http://www.well.com/~art/suicidenotes.html?w}
We will continue to add more data to these corpora as we find them, and all three corpora are freely available.

% Our basic visualizations quantify the ratio of polar words and the ratio of emotion words in the target text. 
% All figures shown in this paper are generated by the emotion analyzer.\footnote{

\begin{figure}[t]
  \begin{center}
  \includegraphics[width=\columnwidth]{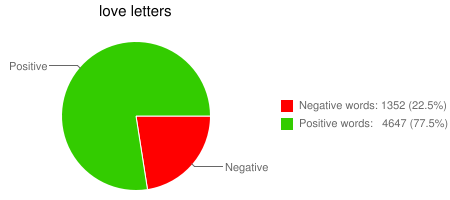}
  \end{center}
  \caption{Percentage of positive and negative words in the love letters corpus.}
  \label{fig:llc-polpie}
  \end{figure}

\begin{figure}[t]
  \begin{center}
  \includegraphics[width=\columnwidth]{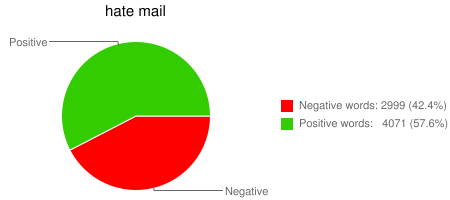}
  \end{center}
  \caption{Percentage of positive and negative words in the hate mail corpus.}
  \label{fig:hmc-polpie}
  \end{figure}

\begin{figure}[t]
  \begin{center}
  \includegraphics[width=\columnwidth]{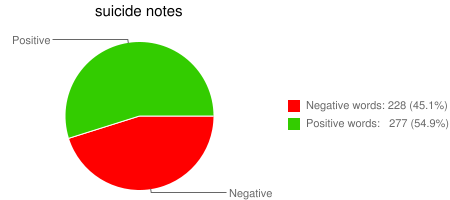}
  \end{center}
  \caption{Percentage of positive and negative words in the suicide notes corpus.}
  \label{fig:snc-polpie}
  \end{figure}

\begin{figure}[t]
  \begin{center}
  \includegraphics[width=\columnwidth]{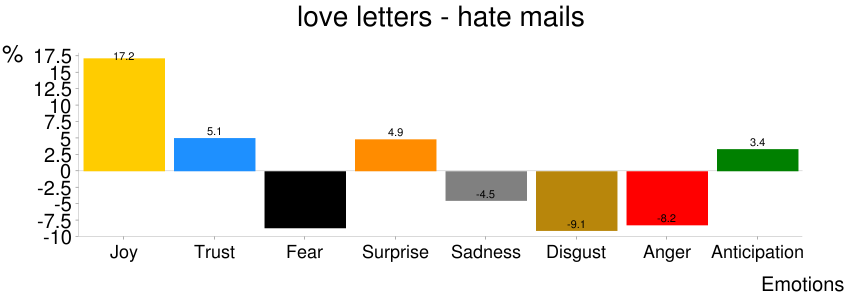}
  \end{center}
  \caption{Difference in percentages of emotion words in the love letters corpus and the hate mail corpus.
The relative-salience word cloud for the joy bar is shown in the figure to the right (Figure 8).}
  \label{fig:love-hate-dev}
  \end{figure}

\begin{figure}[t]
  \begin{center}
  \includegraphics[width=\columnwidth]{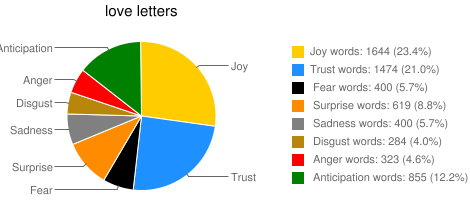}
  \end{center}
  \caption{Percentage of emotion words in the love letters corpus.}
  \label{fig:llc-emopie}
  \end{figure}

\begin{figure}[t]
  \begin{center}
  \includegraphics[width=\columnwidth]{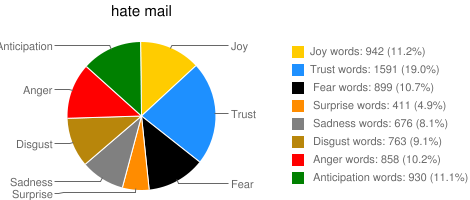}
  \end{center}
  \caption{Percentage of emotion words in the hate mail corpus.}
  \label{fig:hmc-emopie}
  \end{figure}

\begin{figure}[t]
  \begin{center}
  \includegraphics[width=\columnwidth]{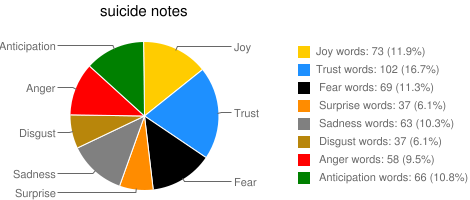}
  \end{center}
  \caption{Percentage of emotion words in the suicide notes corpus.}
  \label{fig:snc-emopie}
  \end{figure}

Figures \ref{fig:llc-polpie}, \ref{fig:hmc-polpie}, and \ref{fig:snc-polpie} show the
percentages of positive and negative words in the love letters corpus, hate mail corpus, and the
suicide notes corpus. % \footnote{Since color is a vital component
% in visualization, the charts are best viewed in a color print out or on screen.}
Figures \ref{fig:llc-emopie}, \ref{fig:hmc-emopie}, and \ref{fig:snc-emopie} show the
percentages of different emotion words in the three corpora.
Emotions are represented by colours as per a study on word--colour associations \cite{Mohammad11a}.
Figure \ref{fig:love-hate-dev} is a bar graph showing the difference of emotion
percentages in love letters and hate mail.
Observe that as expected, love letters have many more joy and trust words, whereas
hate mail has many more fear, sadness, disgust, and anger.

\begin{figure}[t]
  \begin{center}
  \includegraphics[width=\columnwidth]{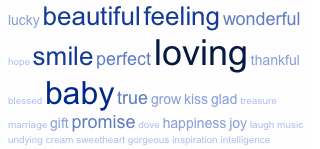}
  \end{center}
  \vspace*{-5mm}
  \caption{Love letters corpus - hate mail corpus: relative-salience word cloud for {\bf joy}.}
  \label{fig:love-hate-joy-cloud}
  \end{figure}

The bar graph is effective at conveying the extent to which one emotion is more prominent in one
text than another, but it does not convey the source of these emotions.
% In order to get a better picture of the emotion words contributing to the salience of an emotion,
Therefore, % in addition to these visualizations, 
we calculate the {\it relative salience} of an emotion word $w$ 
across two target texts $T_1$ and $T_2$:
\vspace*{-2mm}
\begin{equation}
\label{eq:salience}
\text{RelativeSalience}(w|T_1,T_2) = \frac{f_1}{N_1} - \frac{f_2}{N_2} 
\end{equation}
\noindent Where, $f_1$ and $f_2$ are the frequencies of $w$ in $T_1$ and $T_2$, respectively.
$N_1$ and $N_2$ are the total number of word tokens in $T_1$ and $T_2$.
Figure \ref{fig:love-hate-joy-cloud} depicts a relative-salience word cloud of joy words in the love letters 
corpus with respect to the hate mail corpus. 
As expected, love letters, much more than hate mail,
have terms such as {\it loving, baby, beautiful, feeling,} and {\it smile}.
This is a nice sanity check of the manually created emotion
lexicon.  
% on the other hand shows that the salience word cloud of emotion terms helps us quickly interpret emotions in large text files. 
We used Google's freely available software 
to create the word clouds shown in this paper.\footnote{Google WordCloud: http://visapi-gadgets.googlecode.com\\/svn/trunk/wordcloud/doc.html}

\clearpage

% In the subsections below, we describe the various charts shown in the visualization area, followed by additional notes on the text display area. %  when input a target text.
% \subsection{Visualizations}
% \subsubsection{Basic Emotion Analysis}

%  \begin{figure}[]
%  \begin{center}
%  \includegraphics[width=0.6\columnwidth]{LoveHateSuicide/love-hate-wordcloud-disgust.png}
%  \end{center}
%  \caption{Love letters - hate mails: word cloud for disgust.}
%  \label{fig:love-hate-joy-cloud}
%  \end{figure}

\clearpage

  \begin{figure}[t]
  \begin{center}
  \includegraphics[width=\columnwidth]{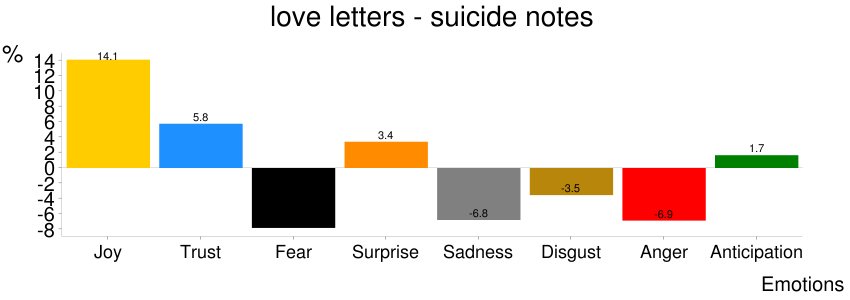}
  \end{center}
  \caption{Difference in percentages of emotion words in the love letters corpus and the suicide notes corpus.}
  \label{fig:love-suicide-dev}
  \end{figure}

%  \begin{figure}[]
%  \begin{center}
%  \includegraphics[width=\columnwidth]{LoveHateSuicide/love-suicide-wordcloud-fear.png}
%  \end{center}
%  \caption{Love letters - suicide notes: word cloud for fear.}
%  \label{fig:love-hate-joy-cloud}
%  \end{figure}

  \begin{figure}[t]
  \begin{center}
  \includegraphics[width=\columnwidth]{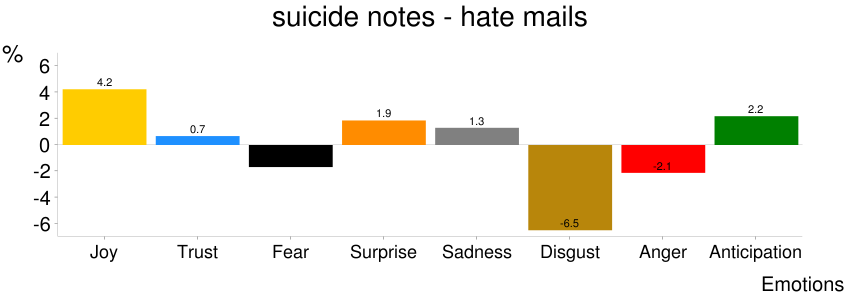}
  \end{center}
  \caption{Difference in percentages of emotion words in the suicide notes corpus and the hate mail corpus.}
  \label{fig:suicide-hate-dev}
  \end{figure}

  \begin{figure}[]
  \begin{center}
  \includegraphics[width=\columnwidth]{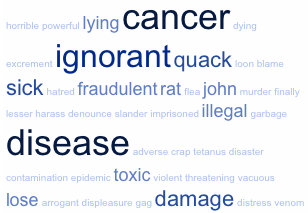}
  \end{center}
  \caption{Suicide notes - hate mail: relative-salience word cloud for {\bf disgust}.}
  \label{fig:suicide-hate-disgust-cloud}
  \end{figure}

Figure \ref{fig:love-suicide-dev} is a bar graph showing the difference
in percentages of emotion words in love letters and suicide notes.
The most salient fear words in the suicide notes with respect to love letters, in decreasing order, were: {\it hell, kill, broke, worship, sorrow, afraid,
loneliness, endless, shaking,} and {\it devil}.

Figure \ref{fig:suicide-hate-dev} is a similar bar graph, but for
suicide notes and hate mail.
Figure \ref{fig:suicide-hate-disgust-cloud} depicts a relative-salience word cloud of disgust words in the hate mail corpus
with respect to the suicide notes corpus.
The cloud shows many words that seem expected, for example {\it ignorant, quack, fraudulent, illegal, lying,} and
{\it damage}. Words such as {\it cancer} and {\it disease} are prominent in this hate mail
corpus because the {\it Millenium Project} denigrates various alternative treatment websites
for cancer and other diseases, and consequently receives angry emails from some cancer patients and physicians.

\section{Emotions in email: men vs. women}

% \cite{Boneva} conducted thorough manual examinations of emails sent by a few hundred households
% in Pittsburgh during a few months spanning 1995, 1996, and 1998. They also conducted several
% interviews of the subjects, and made several claims such as:
There is a large amount of work at the intersection of gender and language (see bibliographies compiled by Schiffman \shortcite{Schiffman02} and Sunderland et al.\@ \shortcite{SunderlandDB02}).
It is widely believed that men and women use language differently, and this is true even in computer mediated communications such as email \cite{Boneva01}.
It is claimed that women tend to foster personal relations \cite{Deaux87,Eagly84} whereas men communicate for social position \cite{Tannen92}.
Women tend to share concerns and support others \cite{Boneva01} whereas men prefer to talk about activities 
\cite{Caldwell82,Davidson82}.
% Duck & Wright, 1993; Spence & Buckner, 1995; Twenge, 1997; Walker, 1994; Wright & Scanlon, 1991),
There are also claims that men tend to be more confrontational and challenging than women.\footnote{http://www.telegraph.co.uk/news/newstopics/\\howaboutthat/6272105/Why-men-write-short-email-and-women-write-emotional-messages.html}

Otterbacher \shortcite{Otterbacher10} investigated stylistic differences in how men and women write product reviews.
Thelwall et al.\@ \shortcite{Thelwall10} examine how men and women communicate over social networks such as MySpace.
Here, for the first time using an emotion lexicon of more than 14,000 words, we investigate if there are gender differences in the use of emotion words
in work-place communications, and if so, whether they support some of the claims mentioned in the above paragraph.
% There are several theories on how men and women are different
% in terms of how strongly they express different emotions in inter-personal communications.
% For example, 
% communications sent by women are on average more cheerful than that sent by men,
% men tend to be more confrontational and challenging than women.
The analysis shown here, does not prove the propositions; however, it provides empirical
support to the claim that men and women use emotion words to different degrees.
% only to determine if the use of emotion words by men and women is consistent with the above theories.
% Gender and language \cite{Schiffman02}
% \cite{SunderlandDB02}
% 
% Gender and email: 
% \cite{Thomson01}: There are clues in email that can be used to identify gender of sender: humans can detect this.
% \cite{Corney02}: men do "report talk", women "rapport talk" 
% -- surface features like wordlength and stylemarkers

We chose the Enron email corpus \cite{Enron}\footnote{The Enron email corpus
is available at http://www-2.cs.cmu.edu/~enron}
as the source of work-place communications because it remains the only large publicly available collection of emails.
It consists of more than 200,000 emails sent between October 1998 and June 2002 by 150 people in senior managerial positions 
at the Enron Corporation, a former American energy, commodities, and services company.
The emails largely pertain to official business but also contain 
personal communication.

  \begin{figure}[t]
  \begin{center}
  \includegraphics[width=\columnwidth]{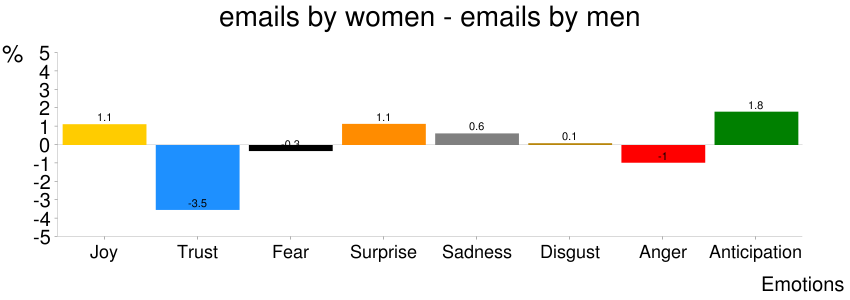}
  \end{center}
  \caption{Difference in percentages of emotion words in emails sent by women and emails sent by men.}
  \label{fig:female-male-dev}
  \end{figure}

  \begin{figure}[t]
  \begin{center}
  \includegraphics[width=\columnwidth]{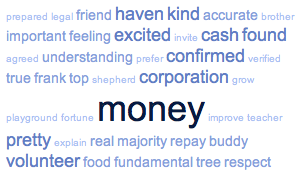}
  \end{center}
  \caption{Emails by women - emails by men: relative-salience word cloud of {\bf trust}.}
  \label{fig:female-male-wordcloud-trust}
  \end{figure}

In addition to the body of the email, the corpus provides meta-information such as the time stamp and the email addresses of the sender and receiver.
Just as in Cheng et al.\@ \shortcite{EnronGender}, (1) we removed emails whose body had fewer than 50 words or more than 200 words,
% so as to filter out long forwards, and very short communications.
%This is consistent with other work on analyzing the Enron corpus \cite{EnronGender}.
% As was done in Cheng et al.\@ \shortcite{EnronGender}, 
(2) the authors manually identified the gender of each of the 150 people solely from their names.
If the name was not a clear indicator of gender, then the person was marked as ``gender-untagged".
This process resulted in tagging 41 employees as female and 89 as male; 20 were left gender-untagged.
Emails sent from and to gender-untagged employees were removed from all further analysis,
leaving 32,045 mails (19,920 mails sent by men and 12,125 mails sent by women).
We then determined the number of emotion words in emails written by men, in emails written by women,
in emails written by men to women, men to men, women to men, and women to women.

  \begin{figure}[t]
  \begin{center}
  \includegraphics[width=\columnwidth]{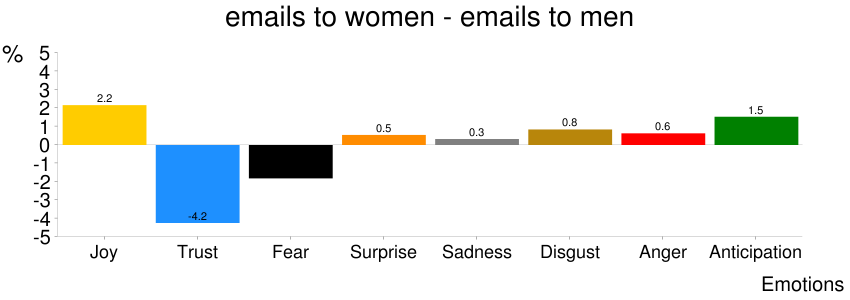}
  \end{center}
  \caption{Difference in percentages of emotion words in emails sent to women and emails sent to men.}
  \label{fig:towomen-tomen-bar-dev}
  \end{figure}

  \begin{figure}[t]
  \begin{center}
  \includegraphics[width=\columnwidth]{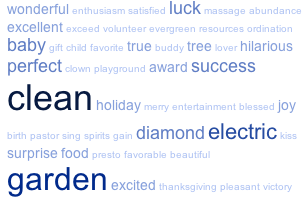}
  \end{center}
  \vspace*{-2mm}
  \caption{Emails to women - emails to men: relative-salience word cloud of {\bf joy}.}
  \vspace*{-4mm}
  \label{fig:towomen-tomen-wordcloud-joy}
  \end{figure}

%  \clearpage 

  \begin{figure}[t]
  \begin{center}
  \includegraphics[width=\columnwidth]{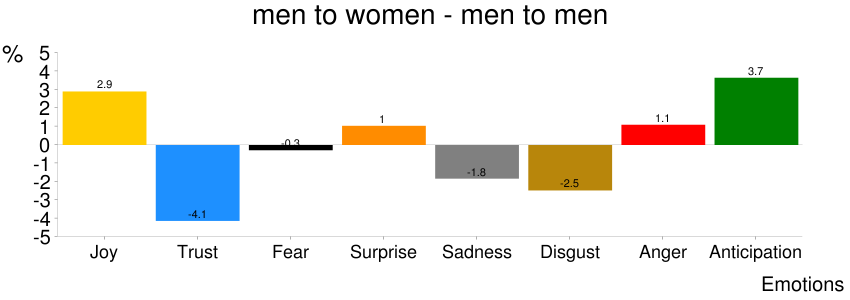}
  \end{center}
  \vspace*{-5mm}
  \caption{Difference in percentages of emotion words in emails sent by men to women and by men to men.}
  \vspace*{-2mm}
  \label{fig:men2women-men2men-bar-dev}
  \end{figure}

  \begin{figure}[t]
  \begin{center}
  \includegraphics[width=\columnwidth]{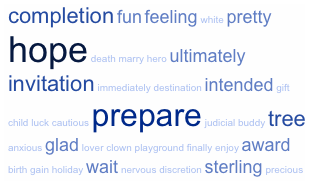}
  \end{center}
  \vspace*{-2mm}
  \caption{Emails by men to women - email by men to men: relative-salience word cloud of {\bf anticipation}.}
  \vspace*{-2mm}
  \label{fig:men2women-men2men-wordcloud-anticip}
  \end{figure}

 \subsection{Analysis}

Figure \ref{fig:female-male-dev} shows the difference
in percentages of emotion words in emails
sent by men from the percentage of emotion words in emails sent by women. Observe the marked difference is in the percentage of trust words.
The men used many more trust words than women.
Figure \ref{fig:female-male-wordcloud-trust} shows the relative-salience word cloud of these trust words.

Figure \ref{fig:towomen-tomen-bar-dev} shows the difference
in percentages of emotion words in emails
sent {\it to} women and the percentage of emotion words in emails sent {\it to} men. Observe the marked difference once again in the percentage of trust words
and joy words.
The men receive emails with more trust words, whereas women receive emails with more joy words.
Figure \ref{fig:towomen-tomen-wordcloud-joy} shows the relative-salience word cloud of joy.

Figure \ref{fig:men2women-men2men-bar-dev} shows the difference in emotion words in emails
sent by men to women and the emotions in mails sent by men to men. Apart from trust words, there is a marked difference in the percentage of anticipation words.
The men used many more anticipation words when writing to women, than when writing to other men.
Figure \ref{fig:men2women-men2men-wordcloud-anticip} shows the relative-salience word cloud of these anticipation words.
% Figures \ref{fig:women2women-women2men-bar-dev}, \ref{fig:women2women-women2men-wordcloud-joy}, and \ref{fig:women2women-women2men-wordcloud-sadness}
% show how women deviate when they write to men as compared to other women.

\begin{figure}[t]
  \begin{center}
  \includegraphics[width=\columnwidth]{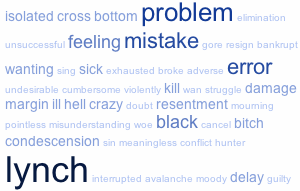}
  \end{center}
  \vspace*{-2mm}
  \caption{Emails by women to women - emails by women to men: relative-salience word cloud of {\bf sadness}.}
  \vspace*{-2mm}
  \label{fig:women2women-women2men-wordcloud-sadness}
  \end{figure}

\begin{figure}[t]
  \begin{center}
  \includegraphics[width=\columnwidth]{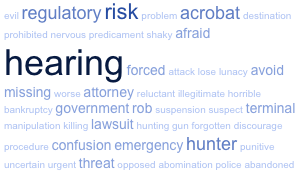}
  \end{center}
   \vspace*{-2mm}
  \caption{Emails by men to men - emails by women to women: relative-salience word cloud of {\bf fear}.}
   \vspace*{-2mm}
  \label{fig:men2men-women2women-wordcloud-fear}
  \end{figure}

Figures \ref{fig:women2women-women2men-bar-dev},
\ref{fig:women2women-women2men-wordcloud-sadness},
\ref{fig:men2men-women2women-bar-dev}, and
\ref{fig:men2men-women2women-wordcloud-fear} show difference bar graphs and relative-salience word clouds
analyzing some other possible pairs of correspondences.

% Figure \ref{fig:women2women-women2men-bar-dev} shows the difference in emotion words in emails
% sent by women to women and the emotion words in mails sent by women to men. 
% Observe that women used lots more joy and sadness words when writing to other women, than when writing to men.
% % Figure \ref{fig:women2women-women2men-wordcloud-joy} and 
% Figure \ref{fig:women2women-women2men-wordcloud-sadness} 
% shows the relative-salience word cloud of the sadness words.
% 
% Figure \ref{fig:men2men-women2women-bar-dev} shows the difference in emotion words in emails
% sent by men to men and the emotion words in mails sent by women to women. 
% Observe that men used lots more fear words when writing to other men, than women do when writing to other women.
% Figure \ref{fig:men2men-women2women-wordcloud-fear} shows the relative-salience word cloud of these fear words.

\subsection{Discussion}
\begin{figure}[t]
  \begin{center}
  \includegraphics[width=\columnwidth]{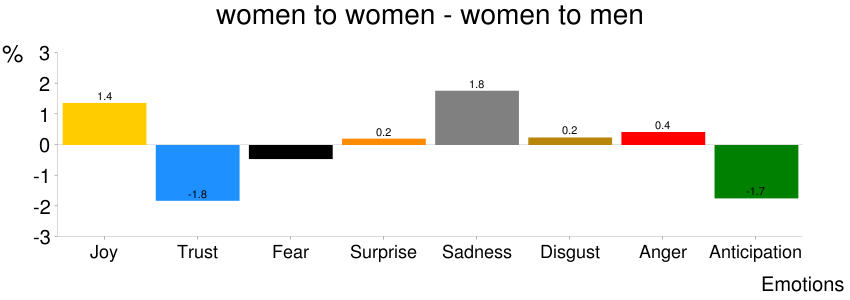}
  \end{center}
  \vspace*{-5mm}
  \caption{Difference in percentages of emotion words in emails sent by women to women and by women to men.}
  \vspace*{-2mm}
  \label{fig:women2women-women2men-bar-dev}
  \end{figure}

\begin{figure}[t]
  \begin{center}
  \includegraphics[width=\columnwidth]{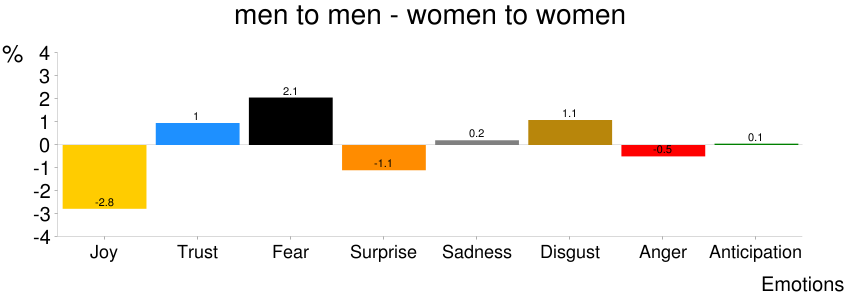}
    \end{center}
	\vspace*{-2mm}
  \caption{Difference in percentages of emotion words in emails sent by men to men and by women to women.}
  \vspace*{-2mm}
  \label{fig:men2men-women2women-bar-dev}
  \end{figure}

Figures \ref{fig:towomen-tomen-bar-dev}, \ref{fig:men2women-men2men-bar-dev}, \ref{fig:women2women-women2men-bar-dev}, and \ref{fig:men2men-women2women-bar-dev}
support the claim that when writing to women, both men and women use more joyous and cheerful
words than when writing to men.
Figures  \ref{fig:towomen-tomen-bar-dev}, \ref{fig:men2women-men2men-bar-dev} and \ref{fig:women2women-women2men-bar-dev} 
show that both men and women use lots of trust words when writing to men.
Figures \ref{fig:female-male-dev}, \ref{fig:women2women-women2men-bar-dev}, and \ref{fig:men2men-women2women-bar-dev} are consistent with the notion that women 
use more cheerful words in emails than men. The sadness values in these figures are consistent with the claim that
women tend to share their worries with other women more often than men with other men, men with women, and women with men.
The fear values in the Figures \ref{fig:men2women-men2men-bar-dev} and \ref{fig:men2men-women2women-bar-dev} suggest that men prefer to use a lot of fear words, especially when communicating with other men.
Thus, women communicate relatively more on the joy--sadness axis,
whereas men have a preference for the trust--fear axis.
It is interesting how 
there is a markedly higher percentage of anticipation
words in cross-gender communication 
than in same-sex communication (Figures \ref{fig:men2women-men2men-bar-dev}, \ref{fig:women2women-women2men-bar-dev}, and \ref{fig:men2men-women2women-bar-dev}). %, and this merits further investigation.

% This is consistent with the claim that men tend to be more confrontational.
% Proving the above claims, however, will require more sophisticated sentence- and discourse-level analyses.

%  \begin{figure}[t]
%  \begin{center}
%  \includegraphics[width=\columnwidth]{MaleFemale/women2women-women2men-wordcloud-joy.png}
%  \end{center}
%  \caption{women to women - women to men: joy words.}
%  \label{fig:women2women-women2men-wordcloud-joy}
%  \end{figure}

\section{Tracking Sentiment in Personal Email}
In the previous section, we showed analyses of sets of emails that were
sent across a network of individuals. In this section, we show visualizations
catered toward individuals---who in most cases have
access to only the emails they send and receive.
% The graphs we will show are useful not only to analyze inter-personal relations
% but also for tracking ones own emotions. 
We are using Google Apps API to develop an application that integrates with Gmail (Google's email service),
to provide users with the ability to track their emotions towards people
they correspond with.\footnote{Google Apps API: http://code.google.com/googleapps/docs} 
Below we show some of these visualizations
by selecting John Arnold, a former employee at Enron, as a stand-in
for the actual user.

\begin{figure}[h]
  \begin{center}
  \includegraphics[width=\columnwidth]{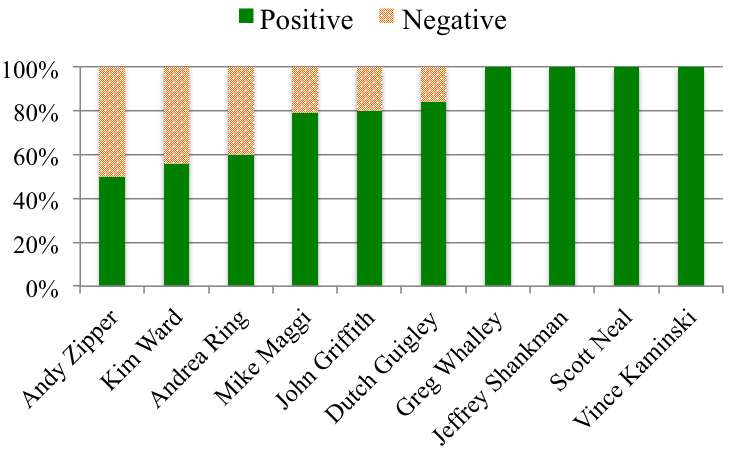}
  \end{center}
   \vspace*{-4mm}
  \caption{Emails sent by John Arnold.}
   \vspace*{-1mm}
  \label{fig:johnemails}
  \end{figure}

\begin{figure}[t]
  \begin{center}
  \includegraphics[width=\columnwidth]{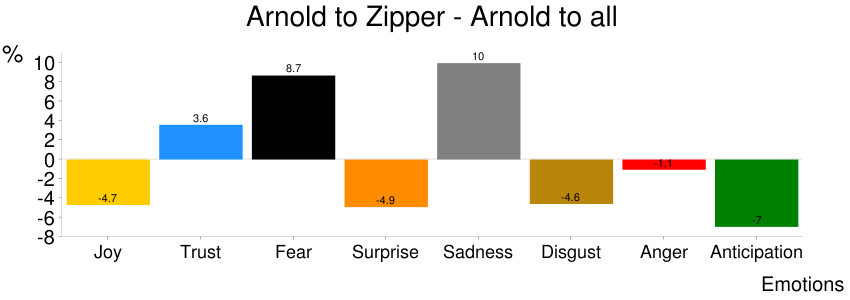}
  \end{center}
   \vspace*{-2mm}
  \caption{Difference in percentages of emotion words in emails sent by John Arnold to Andy Zipper and emails sent by John to all.}
   \vspace*{-4mm}
  \label{fig:andy-zipper-bar-dev}
  \end{figure}

\begin{figure}[t]
  \begin{center}
  \includegraphics[width=\columnwidth]{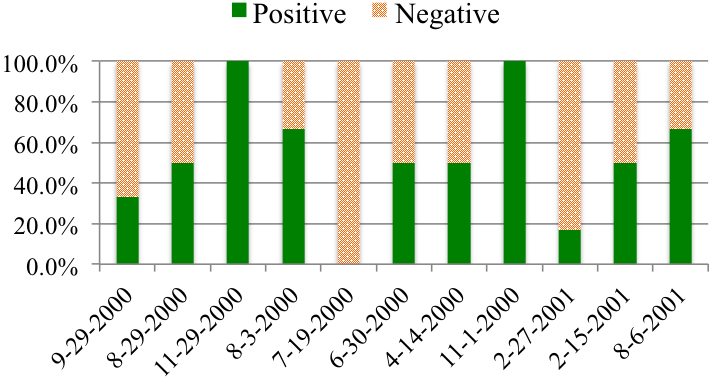}
  \end{center}
   \vspace*{-3mm}
  \caption{Emails sent by John Arnold to Andy Zipper.}
  \label{fig:johntoandy}
  \end{figure}

Figure \ref{fig:johnemails} shows the percentage of positive and negative
words in emails sent by John Arnold
to his colleagues. 
John can select any of the bars in the figure to reveal the difference in percentages of emotion words in emails sent
to that particular person and all the emails sent out.
Figure \ref{fig:andy-zipper-bar-dev} shows the graph
pertaining to Andy Zipper.
Figure \ref{fig:johntoandy} shows the percentage of positive and negative words
in each of the emails sent by John to Andy.
% The user can also view relative-salience word clouds for different emotions.

In the future, we will make a public call for volunteers interested in our Gmail
emotion application, and we will request  access to numbers of emotion words in their emails
for a large-scale analysis of emotion words in personal email.
The application will protect the privacy of the users by passing 
emotion word frequencies, gender, and age, but no text, names, or email ids.

\section{Conclusions}
We have created a large word--emotion association lexicon by crowdsourcing,
and used it to analyze and track the distribution of emotion words in mail.\footnote{Please send an e-mail to
saif.mohammad@nrc-cnrc.gc.ca to obtain the latest version of the NRC Emotion Lexicon, suicide notes corpus, hate mail corpus,
love letters corpus, or the Enron gender-specific emails.}
We compared emotion words in love letters, hate mail,
and suicide notes. 
We analyzed 
% the difference in the amount of emotion words
% used by men and women in 
work-place email
and showed that women use and receive a relatively larger number of joy and sadness words,
whereas men use and receive a relatively larger number of trust and fear words.
We also found that there is a markedly higher percentage of anticipation
words in cross-gender communication (men to women and women to men)
than in same-sex communication.
We showed how different visualizations and word clouds can be used 
to effectively interpret the results of the emotion analysis.
Finally, we showed additional 
visualizations and a Gmail application that can help people track emotion words in the
emails they send and receive.

 % The system aslo shows tag clouds of various emotion terms.

% \section*{Acknowledgments}
% We thank the reviewers for their time and comments.

% \vspace*{-1mm}
% {\small
\section*{Acknowledgments}
\vspace*{-2mm}
Grateful thanks to Peter Turney and Tara Small for many wonderful ideas.
Thanks to the thousands of people who answered the emotion survey with diligence and care.
This research was funded by National Research Council Canada.
%  }

\clearpage

\bibliography{references}

%   \begin{figure}[t]
%   \begin{center}
%   \includegraphics[width=\columnwidth]{manwoman-anger-timeline.png}
%   \end{center}
%   \caption{Percentage of {\bf anger} words in close proximity to occurrences of {\it man} and {\it woman} in books from the year 1800 to 2004.
% Source: 5-gram data released by Google.}
%   \label{fig:manwoman-anger-timeline}
%   \end{figure}

\end{document}